\documentclass[10pt, conference, a4paper]{IEEEtran}
 \usepackage{cite}

\ifCLASSINFOpdf
\usepackage[pdftex]{graphicx}
\else
\fi

\usepackage{amsmath}

\usepackage{algorithmic}

\usepackage{array}

\usepackage{float}

\usepackage{url}

\begin{document}

\title{Regularized Dynamic Boltzmann Machine with Delay Pruning for Unsupervised Learning of Temporal Sequences}

\author{\IEEEauthorblockN{Sakyasingha Dasgupta}
\IEEEauthorblockA{IBM Research - Tokyo\\
Email: sdasgup@jp.ibm.com}
\and
\IEEEauthorblockN{Takayuki Yoshizumi }
\IEEEauthorblockA{IBM Research - Tokyo\\
Email: yszm@jp.ibm.com}
\and
\IEEEauthorblockN{Takayuki Osogami}
\IEEEauthorblockA{IBM Research - Tokyo\\
Email: osogami@jp.ibm.com}}


%


\maketitle

\begin{abstract}
We introduce Delay Pruning, a simple yet powerful technique to regularize dynamic Boltzmann machines (DyBM). The recently introduced DyBM provides a particularly structured Boltzmann machine, as a generative model of a multi-dimensional time-series. This Boltzmann machine can have infinitely many layers of units but allows exact inference and learning based on its biologically motivated structure. DyBM uses the idea of conduction delays in the form of fixed length first-in first-out (FIFO) queues, with a neuron connected to another via this FIFO queue, and spikes from a pre-synaptic neuron travel along the queue to the post-synaptic neuron with a constant period of delay. Here, we present Delay Pruning as a mechanism to prune the lengths of the FIFO queues (making them zero) by setting some delay lengths to one with a fixed probability, and finally selecting the best performing model with fixed delays. The uniqueness of structure and a non-sampling based learning rule in DyBM, make the application of previously proposed regularization techniques like Dropout or DropConnect difficult, leading to poor generalization. First, we evaluate the performance of Delay Pruning to let DyBM learn a multidimensional temporal sequence generated by a Markov chain. Finally, we show the effectiveness of delay pruning in learning high dimensional sequences using the moving MNIST dataset, and compare it with Dropout and DropConnect methods. 
\end{abstract}


%
\IEEEpeerreviewmaketitle

\section{Introduction}
Deep neural networks \cite{lecun2015deep}, \cite{salakhutdinov2009deep} have  been successfully applied for learning in a large number of image recognition and other machine learning tasks. However, neural network (NNs) based models are typically well suited on scenarios with large amounts of available labelled datasets. Increasing the network complexity (in terms of size or number of layers), one can achieve impressive levels of performance. A caveat is that this can lead to gross over-fitting or generalization issues, when trained in the presence of limited amount of training samples. As a result, a wide range of techniques, like adding a $L2$ penalty term, Bayesian methods \cite{williams1995bayesian}, adding noise to training data \cite{bishop1995training} etc., for regularizing NNs have been developed. 

More recently, with a focus on NNs with a deep architecture, Dropout \cite{srivastava2014dropout} and DropConnect \cite{wan2013regularization} techniques have been proposed as ways to prevent over-fitting by randomly omitting some of the feature detectors on each training sample. Specifically, Dropout involves  randomly deleting some of the activations (units) in each layer during a forward pass and then back-propagating the error only through the remaining units. DropConnect generalizes this to randomly omitting weights rather than the activations (units). Both these techniques have been shown to significantly improve the performance on standard fully-connected deep neural network architectures.

In this work, we propose a novel regularization technique called Delay Pruning, designed for a recently introduced generative model called dynamic Boltzmann machine (DyBM) \cite{osogami2015seven}. Unlike the conventional Boltzmann machine (BM) \cite{ackley1985learning}, which is trained with a collection of static patterns, DyBM is designed for unsupervised learning of temporal pattern sequences. DyBM is motivated by postulates and observations from biological neural networks, allowing exact inference and learning of weights based on the timing of spikes (spike-timing dependent plasticity - STDP). Unlike the restricted Boltzmann machine (RBM) \cite{salakhutdinov2007restricted}, DyBM has no specific hidden units, and the network can be unfolded through time, allowing infinitely many layers \cite{osogami2015learning}. Furthermore, DyBM can be viewed as fully-connected recurrent neural network with memory units and with conduction delays between units implemented in the form of fixed length first-in first-out (FIFO) queues. A spike originating at a pre-synaptic neuron (unit) travels along this FIFO queue and reaches the post-synaptic neuron after a fixed delay. The length of the FIFO queues is equal to one minus the maximum delay value. Due to this completely novel architecture of DyBM applying existing regularization methods is difficult or does not lead to better generalization performance. 

As such, the here proposed Delay Pruning technique allows a method for regularized training of NNs with FIFO queues. Specifically, during training, it truncates the lengths to zero, for randomly selected FIFO queues. We evaluate the performance of Delay Pruning on a stochastic multi-dimensional time series and then compare it with Dropout and DropConnect for unsupervised learning on the high-dimensional moving MNIST dataset. In the next sections, we first give a brief overview of DyBM and its learning rule, followed by the Delay Pruning algorithm, experimental results and conclusion.

\section{Dynamic Boltzmann Machine}

\subsection{Overview}
In this paper, we use DyBM \cite{osogami2015seven} for unsupervised learning of temporal sequences and show better generalised performance using our Delay Pruning algorithm. Unlike standard Boltzmann machines, DyBM can be trained with a time-series of patterns. Specifically, the DyBM gives the conditional probability of the next values (patterns) of a time-series given its historical values.  This conditional probability can depend on the whole history of the time-series, and the DyBM can thus be used iteratively as a generative model of a time-series. 

DyBM can be defined from BM having multiple layers of units, where one layer represents the most recent values of a time-series, and the remaining layers represent the historical values of the time-series. The most recent values are conditionally independent of each other given the historical values.  The DyBM is equivalent to such a BM having an infinite number of layers, so that the most recent values can depend on the whole history of the time series.  We train the DyBM in such a way that the likelihood of given time-series is maximized with respect to the conditional distribution of the next values given the historical values. Similar to a BM, a DyBM consists of a network of artificial neurons.  Each neuron takes a binary value, 0 or 1, following a probability distribution that depends on the parameters of the DyBM. Unlike the BM, the values of the DyBM can change over
time in a way that depends on its previous values.  That is, the DyBM stochastically generates a multi-dimensional series of binary values.

Learning in conventional BMs is based on an Hebbian formulation, but is often approximated with sampling based strategy like contrastive divergence. In this formulation the concept of time is largely missing. In DyBM, like biological networks, learning is dependent on the timing of spikes. This is called spike-timing dependent plasticity, or STDP \cite{song2001cortical}, which states that a synapse is strengthened if the spike of a pre-synaptic neuron precedes the spike of a post-synaptic neuron (long term potentiation - LTP), and the synapse is weakened if the temporal order is reversed (long term depression - LTD). DyBM uses an exact online learning rule, that has the properties of LTP and LTD. 

\begin{figure}[ht]
\centering
\includegraphics[width= \linewidth]{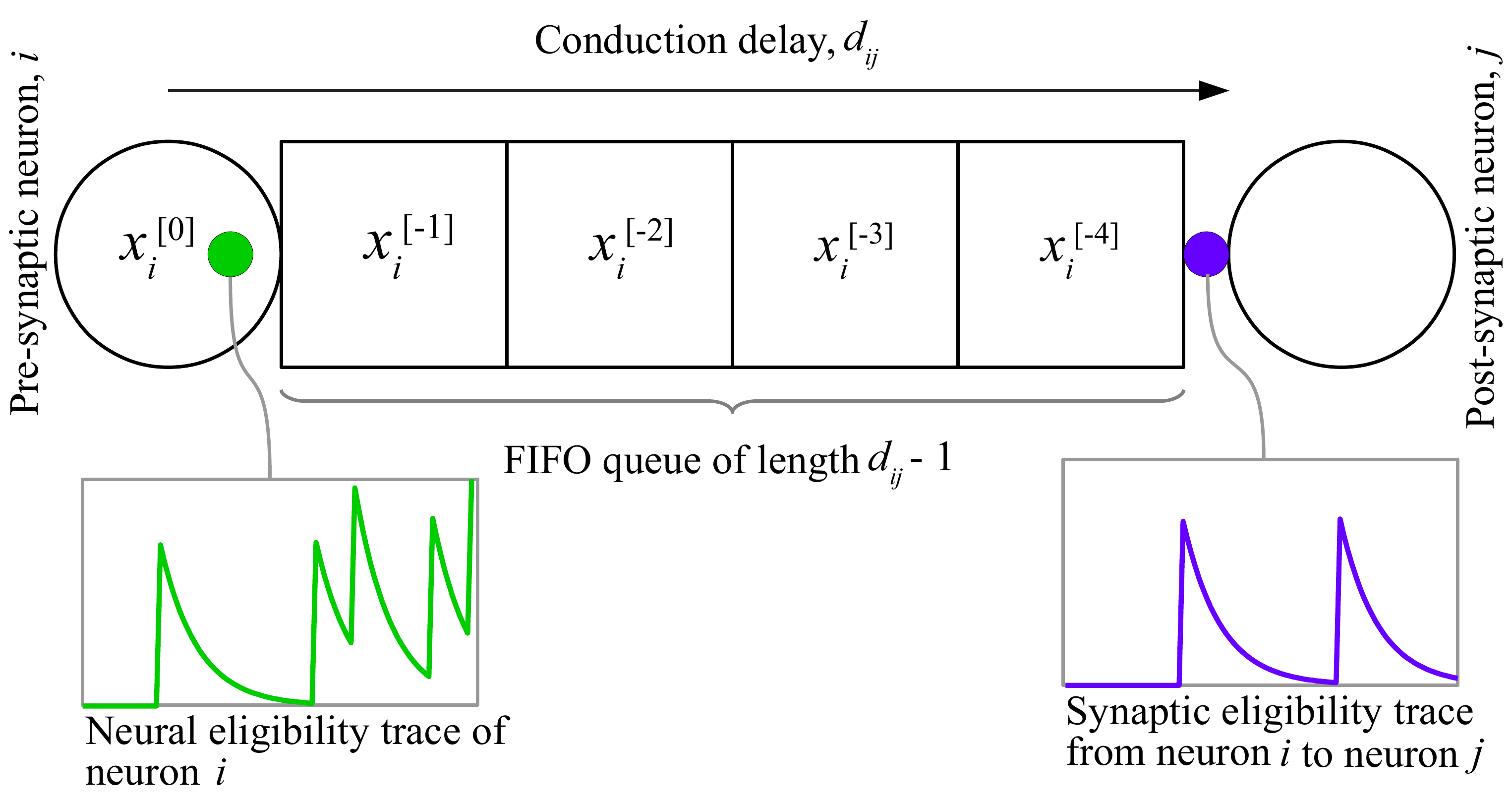}
\caption{
A DyBM consists of a network of neurons and memory units.  A
 pre-synaptic neuron is connected to a post-synaptic neuron via a FIFO
 queue.  The spike from the pre-synaptic neuron reaches the
 post-synaptic neuron after a constant conduction delay.  Each neuron
 has the memory unit for storing neural eligibility traces.  A synaptic eligibility
 trace is associated with a synapse between a pre-synaptic neuron and a
 post-synaptic neuron, and summarizes the spikes that have arrived at
 the synapse, via the FIFO queue. }
 \label{fig:1}
 \end{figure}

The learning rule of DyBM exhibits some of the key properties of STDP due to its structure consisting of conduction delays and memory units, which are illustrated in Figure 1. A neuron is connected to another in a way that a spike from a pre-synaptic neuron, $i$, travels along an axon and reaches a post-synaptic neuron, $j$, via a synapse after a delay consisting of a constant period, $d_{i,j}$.  In the DyBM, a FIFO queue causes this conduction delay.  The FIFO queue stores the values of the pre-synaptic neuron for the last $d_{i,j}-1$ units of time.  Each stored value is pushed one position toward the head of the queue when the time is incremented by one unit.  The value of the pre-synaptic neuron is thus given to the post-synaptic neuron after the conduction delay. Moreover, the DyBM aggregates information about the spikes in the past into neural eligibility traces and synaptic eligibility traces, which are stored in the memory units. Each neuron is associated with a learnable parameter called bias. The strength of the synapse between a pre-synaptic neuron and a post-synaptic neuron is represented by learnable parameters called weights. Those are further divided into LTP and LTD components. 

\subsection{Definition}
\begin{figure}
\begin{minipage}{0.35\linewidth}
\centering
\includegraphics[scale = 0.13]{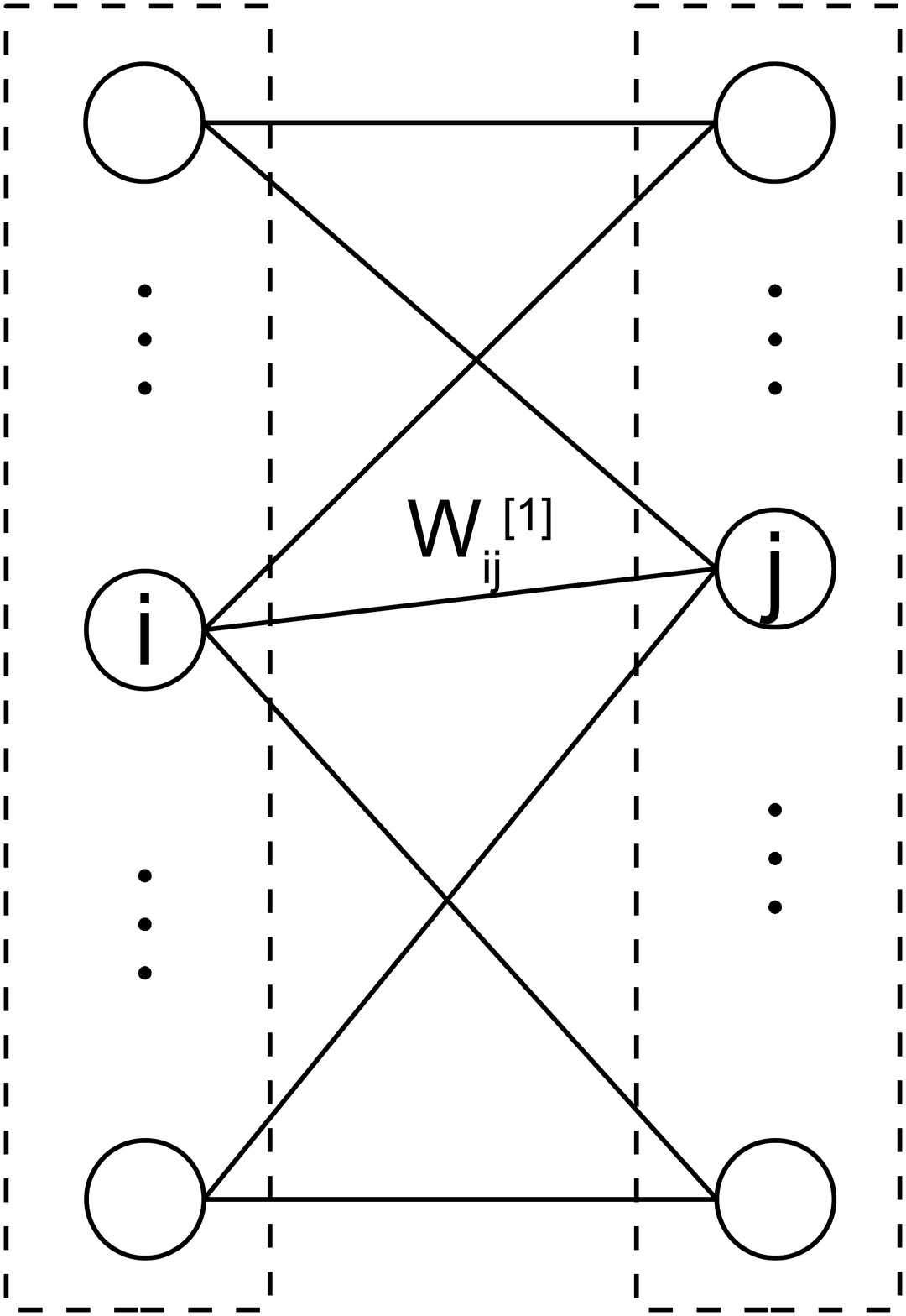}\\
(a) RBM
\end{minipage}
\begin{minipage}{0.44\linewidth}
\centering
\includegraphics[scale = 0.12]{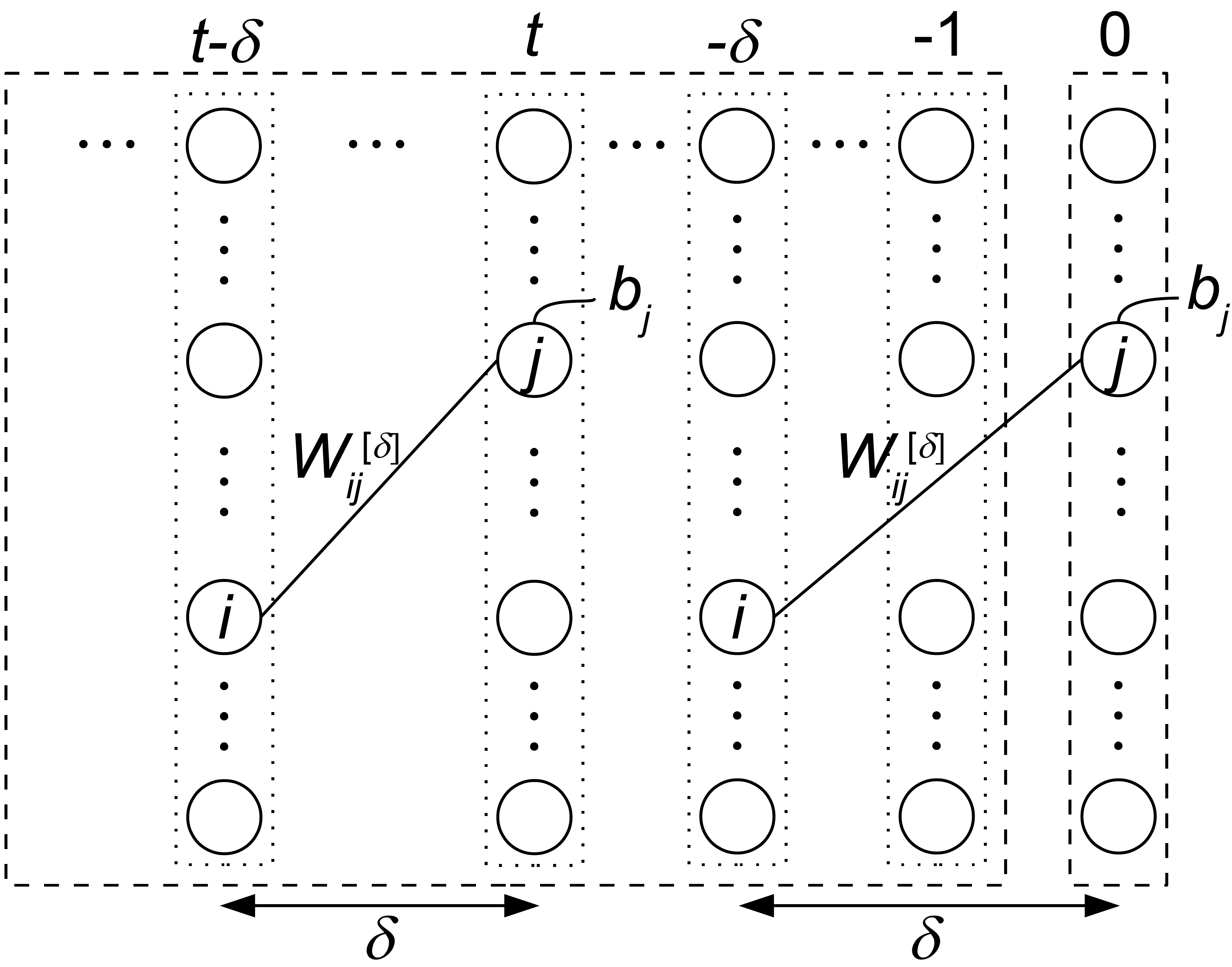}\\
(b) DyBM
\end{minipage}
\caption{(a) A restricted Boltzmann machine, and (b) dynamic Boltzmann machine unfolded in time i.e. $t \rightarrow \infty$ \cite{osogami2015learning}.}
\label{fig:BM}
\end{figure}

The DyBM shown in Figure~\ref{fig:BM}~(b) can be shown to be equivalent to a BM having infinitely many layers of units \cite{osogami2015learning}.  Similar to the RBM (Figure~\ref{fig:BM}~(a)), the DyBM has no weight
between the units in the right-most layer of Figure~\ref{fig:BM}~(b).
Unlike the RBM \cite{salakhutdinov2007restricted}, each layer of the DyBM has a common number, $N$, of
units, and the bias and the weight in the DyBM can be shared among
different units in a particular manner.  

Formally, the DyBM-$T$ is a BM having $T$ layers from $-T+1$ to $0$, where $T$ is a positive integer or infinity.  Let $\mathbf{x}\equiv(\mathbf{x}^{[t]})_{-T<t\le 0}$, where $\mathbf{x}^{[t]}$ is the values of the units in the $t$-th layer, which we consider as the values at time $t$. The $N$ units at the 0-th layer (the right-most layer of Figure~\ref{fig:BM}~(b)) have an associated bias term $\mathbf{b}$.  For any $\delta\ge 1$, $\mathbf{W}^{[\delta]}$ gives the matrix whose $(i,j)$ element, $W_{i,j}^{[\delta]}$, denotes the weight between the $i$-th unit at time $-\delta$ and the $j$-th unit at time $0$ for any $\delta$. This weight can in turn be divided into LTP and LTD components. As introduced in the previous section, each neuron stores a fixed number, $L$, of neural eligibility traces.  For $\ell\in[1,L]$ and $j\in[1,N]$, $\gamma_{j,\ell}^{[t-1]}$ is the $\ell$-th
neural eligibility trace of the $j$-th neuron immediately before time $t$. This is calculated as weighted sum of the past values of that neuron, with recent values weighing more:
\begin{align}
\gamma_{j,\ell}^{[t-1]} \equiv \sum_{s=-\infty}^{t-1} \mu_{\ell}^{t-s}
 \, x_j^{[s]},
 \label{eq:neural_eligibility}
\end{align}
where, $\mu_\ell\in(0,1)$ is the decay rate for the $\ell$-th neural eligibility trace.	
Each neuron also stores synaptic eligibility traces as weighted sum of the values that has reached neuron, $j$, from a pre-synaptic neuron, $i$, after the conduction delay, $d_{i,j}$, with recent values weighing more.    Namely, the post-synaptic neuron $j$ stores a fixed number, $K$, of synaptic eligibility traces.  For $k\in[1,K]$, $\alpha_{i,j,k}^{[t-1]}$ is the $k$-th synaptic eligibility trace of the neuron $j$ for the
pre-synaptic neuron $i$ immediately before time $t$:
\begin{align}
\alpha_{i,j,k}^{[t-1]} \equiv \sum_{s=-\infty}^{t-d_{i,j}}
 \lambda_k^{t-s-d_{i,j}} \, x_i^{[s]},
 \label{eq:synaptic_eligibility}
\end{align}
here, $\lambda_k\in(0,1)$ is the decay rate for the $k$-th synaptic eligibility trace. Both of the eligibility traces are updated locally in time as follows:
\begin{align}
\gamma_{j,\ell}^{[t]} & \leftarrow \mu_\ell \left( \gamma_{j,\ell}^{[t-1]} + x_j^{[t]} \right) \\
\alpha_{i,j,k}^{[t]}   & \leftarrow \lambda_k \left( \alpha_{i,j,k}^{[t-1]} + x_i^{[t-d_{i,j}]} \right)
\end{align}
for $\ell\in[1,L]$ and $k\in[1,K]$, and for neurons $i$ that are connected to $j$.																																																																								

For a DyBM-$T$, $P_\theta(\mathbf{x}^{[0]} |
\mathbf{x}^{(-T,-1]})$, is the conditional probability of
  $\mathbf{x}^{(0)}$ given $\mathbf{x}^{(-T,-1]}$, where we use
    $\mathbf{x}^{I}$ for an interval $I$ such as $(-T,-1]$ to denote
      $(\mathbf{x}^{[t]})_{t\in I}$.  Because the units in the 0-th layer
have no weight with each other, this conditional probability has the property of conditional independence analogous to RBMs. 

DyBM can be seen as a model of a time-series in the following sense. Specifically, given a history $\mathbf{x}^{(-T,-1]}$ of a time-series, the DyBM-$T$ gives the probability of the next values, $\mathbf{x}^{[0]}$ of the time-series with $P_\theta(\mathbf{x}^{[0]} | \mathbf{x}^{(-T,-1]})$.  With a DyBM-$\infty$, the next
values can depend on the whole history of the time-series.  In
principle, the DyBM-$\infty$ can thus model any time-series
possibly
with long-term dependency, as long as the values of the time-series
at a moment is conditionally independent of each other given its values
preceding that moment.  Using the conditional probability
given by a DyBM-$T$, the probability of a sequence,
$\mathbf{x}=\mathbf{x}^{(-L,0]}$, of length $L$ is given by
\begin{align}
p(\mathbf{x}) = \prod_{t=-L+1}^0 P_\theta(\mathbf{x}^{[t]}| \mathbf{x}^{(t-T, t-1]}),
\label{eq:chain}
\end{align}
where we arbitrarily define $\mathbf{x}^{[t]}\equiv\mathbf{0}$ for $t\le -L$. Namely, the values are set to zero if there are no corresponding history. 

The STDP based learning rule for a DyBM-T is derived such that the log-likelihood of a given set ($\cal D$) of time-series is maximised by maximising the sum of the log-likelihood of $\mathbf{x} \in \cal D$. Using Eq. \ref{eq:chain}, the log-likelihood of $\mathbf{x}=\mathbf{x}^{(-L,0]}$ has the following gradient:

\begin{align}
\nabla_\theta \log p(\mathbf{x}) = \sum_{t=-L+1}^0 \nabla_\theta \log P_\theta(\mathbf{x}^{[t]}| \mathbf{x}^{(t-T,t-1]}).
\label{eq:grad}
\end{align}

Typically, the computation of this gradient can be intractable for large $T$, however in DyBM using a specific form of weight sharing \cite{osogami2015seven}, exact and efficient gradient calculation is possible. Specifically, in the limit of $T \to \infty$ using the formulation of neural and synaptic eligibility traces, the parameters $\theta$ of DyBM can be computed exactly using an online stochastic gradient rule that maximizes the log-liklihood of the given set $\cal D$:

\begin{align}
\theta 
\leftarrow \theta + \eta \, \sum_{\mathbf{x}\in{\cal D}} \nabla_\theta \log P_\theta(\mathbf{x}^{[0]}|\mathbf{x}^{(-\infty,-1]}).
\end{align}

Due to space limitations, the weight update rules are not provided here. See \cite{osogami2015seven} for details.

\section{Regularization with Delay Pruning}
Delay Pruning provides a method of training DyBM, and in general neural networks with FIFO queues, with regularization and then choosing the best performing model for improved prediction on test dataset. Specifically it refers to truncating the FIFO queue lengths to zero, by setting their respective delay values to unit length, for randomly selected axons with a probability $p$. Figure 3. displays the difference in architecture of two connected neurons with a FIFO queue, for the original DyBM and a delay pruned version. The procedure is carried out as follows: 

Initialize DyBM parameters, with the delay length ($d_{ij}$) for the FIFO queues connecting neurons $i$ and $j$, selected randomly within a certain range. Here we use $d_{ij} \in [1,7]$. Each neuron is connected to another neuron with two FIFO queues (outgoing and incoming axon) of lengths initialized to $d_{ij}-1$.  Calculate the negative log-likelihood (original negative log-likelihood - ONL) with respect to the true distribution of the temporal-pattern from the training sample.

\begin{figure} 
    \centering
    \includegraphics[width=0.9\linewidth]{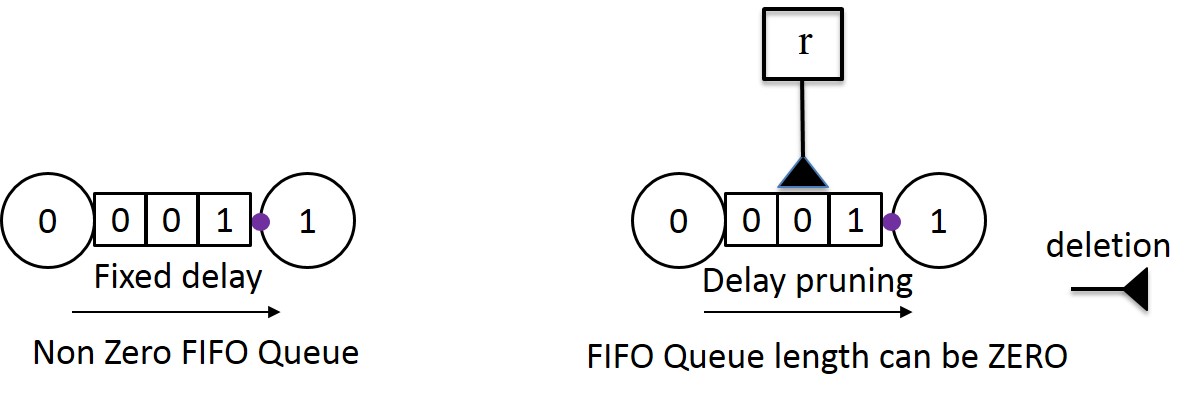}
    \caption{Pictorial representation of Delay Pruning method for dynamic Boltzmann machines. (left) Original DyBM showing a single axonal connection via a FIFO queue between two units.  (right) The setup with delay pruning shown for a single axon.}
    \label{simulationfigure}
\end{figure}

For each training sample and current training cycle:
\begin{itemize}
\item[1] For every fixed number of epochs, validate the previously learned DyBM for predicting a temporal sequence pattern. Calculate negative log-likelihood ( training negative log-likelihood - TNL) of the training (or validation) data with respect to the distribution defined by the trained DyBM. Update the performance evaluation measure $\epsilon$ by calculating the difference between ONL and TNL (any other appropriate performance measure e.g. cross-entropy, can also be used instead). Update a best model pointer to point towards the learnt network with minimum $-\epsilon$ so far.
\item[2] Select random variable ($r \approx \mathtt{Bernoulli}(p)$) from a Bernoulli distribution with probability $p$. If $r=1$, keep the original maximum delay (FIFO Queue length), otherwise, set the current maximum delay $d_{max}=1$. Thus truncating the current FIFO queue length to zero.
\item[3] Repeat till all training cycles are exhausted.
\item[4] The best performing model (with all parameters fixed) from the training and validation process is selected for final testing.
\end{itemize}

Similar to Dropout and DropConnect, applying the delay pruning algorithm amounts to sampling the best performing "thinned" network. In this case the thinned network consists of all FIFO queues that survived the pruning procedure. For each presentation of training cycle, a new thinned network is sampled and trained. As a result of this procedure, one can train across an ensemble of models and thus effectively regularize DyBM to prevent over-fitting. Finally instead of averaging across the ensemble, we select the best performing model. This is analogous to bagging based ensemble learning method in other machine learning areas \cite{hansen1990neural}. 

\section{Experiments}
We designed two different experiments of increasing complexity in order to evaluate the effect of Delay Pruning on DyBM. Given that DyBM is a neural network suitable for learning a generative model of temporal sequences, the two tasks were chosen so as to show the effect of regularization for modelling and predicting high-dimensional temporal patterns. The experiments were conducted using a purely CPU based, Java\textsuperscript{\textregistered}  implementation of the DyBM on a MacBook Air with an Intel Core i5, 1.6 GHz.

\subsection{Training}
DyBM was trained using mini-batches of samples from the training set in both cases. Each sample was trained for a maximum of fifty thousand time steps. The Delay Pruning was carried out continuously. Every $500$ epoch, DyBM was tested on a sample from a validation set (generating a validation temporal pattern sequence), and the currently best performing model was updated. After every mini-batch, all the eligibility traces in DyBM were reinitialized with the learned weights from the previous mini-batch transferred to the next batch. Training stopped if the maximum time was reached or if the  estimated negative log-likelihood of trained DyBM matched that of the true negative log-likelihood of validation set for an entire epoch. The learning rates were initially fixed to a small value and then adjusted during training using the optimisation technique of adaptive moment estimation  \cite{kingma2014adam}, while the parameters of DyBM were learned using a stochastic gradient method.

The bias and weight parameters were initialized randomly from a normal distribution with mean $0.0$ and standard deviation $0.1$. DyBM uses a fully-connected network with each neuron having a self-connection via a FIFO queue. Each neuron held three ($L=3$) neural and three ($K=3$) synaptic eligibility traces, respectively. 

\subsection{Learning multi-dimensional stochastic time-series }
This task involved, learning to model and predict the next sequence of a 7-dimensional stochastic time-series $\mathbf{X_D} = (X_1, X_2, X_3,...,X_7)$, where $X_i \in \{0,1\}^T$. Here, $T$ is the length of time-series. The time-series was synthetically generated using a discrete-time one-Markov process as depicted in Fig. 4(a). The probability to generate the same state of a '0' or '1' was fixed at $P_s(X_i^t) = 0.95$, while the transition probability to a different state from '0' to '1' or vice versa, was fixed at $P_s(X_i^{t+1}=1|X_i^t=0) = 0.05$. The number of training and testing data were set as $100,000$ and $10,000$, respectively. A small training set was deliberately chosen in order to check for the generalisation ability of the original DyBM as compared with the DyBM regularised with Delay Pruning during training. Each FIFO queue connection delay was initialized randomly as $d_{ij} \in [1,7]$. The number of neurons $N = 7$, with each neuron encoding one of the $7$ input dimensions. The probability of pruning was fixed at $p = 0.5$. Fig. 4 (b) shows an example plot with a section of the training dataset. 

\begin{figure*} 
 \centering
 \includegraphics[scale = 0.35]{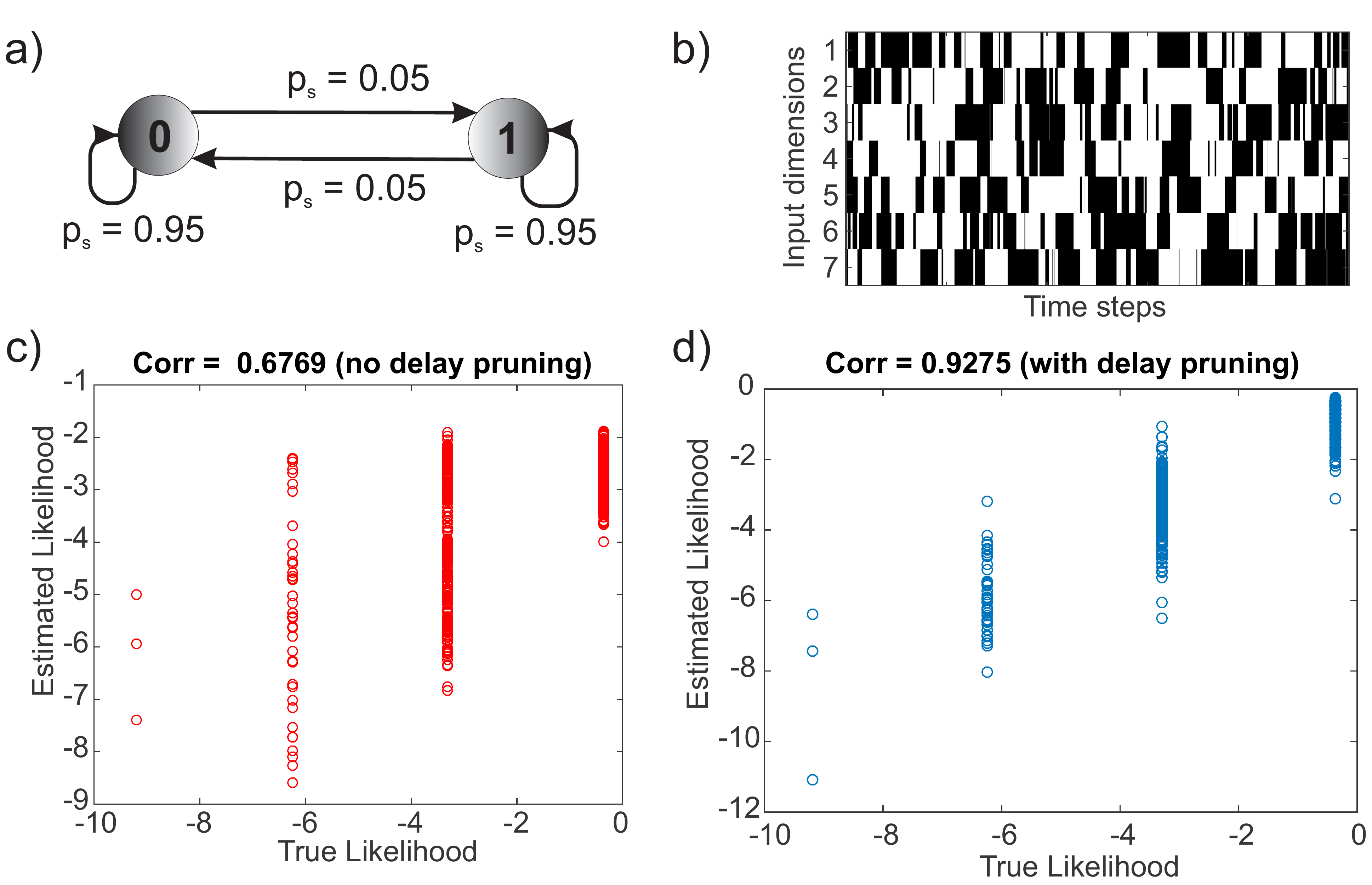}
    \caption{Predicting a multidimensional discrete-time Markov process: (a) state diagram of the one-Markov process (b) a training sample example showing the 7-dimensional stochastic binary time-series. White regions represent '0' and black regions represent '1'; Correlation between negative log-likelihood of predicted series and true negative log-likelihood, for the test data (c) without delay pruning and (d) with delay pruning.}
    \label{simulationfigure}
\end{figure*}

We first trained and tested with the original DyBM without any regularization. In Fig. 4(c), we plot the negative log-likelihood with respect to the true data distribution against the estimated negative log-likelihood of the test data with respect to the distribution defined by the trained DyBM. As observed, it achieved a poor generalization with a low correlation coefficient of $0.6769$. Keeping all parameters the same, re-training DyBM with Delay Pruning regularization method (as explained in Section. 3) resulted in significantly better generalization in the prediction of the test data time-series. This is clearly observed from the high correlation coefficient of $0.9275$ between the estimated negative log-likelihood and the true negative log-likelihood of the data distribution.

\subsection{Moving MNIST Prediction}
Unsupervised learning of image sequences \cite{srivastava2015unsupervised} is a difficult problem. Avoiding over-fitting in order to predict future pattern sequences is considerably challenging. As such, this task  was designed so as to test the ability of delay pruned DyBM to go through a temporal sequence of image frames and learn the underlying representation. We then test it for generating the original input sequences and also for predicting future image frames in the correct temporal order.
 \begin{figure*} 
    \centering
    \includegraphics[scale =0.38]{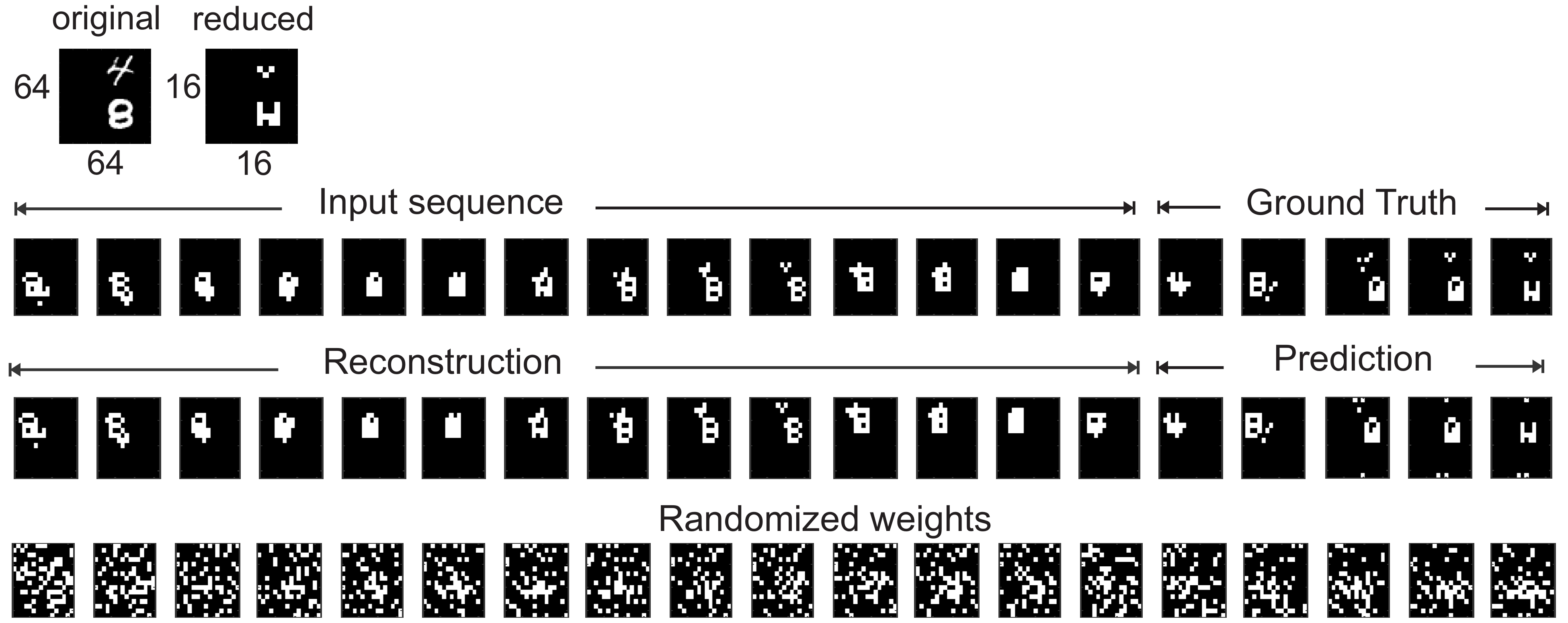}
    \caption{Temporal sequence reconstruction and prediction on the $16 \times 16$ bits reduced resolution moving MNIST dataset. DyBM was trained with the first 14 frames. This is depicted as the input sequence. During testing, DyBM had to reconstruct the original input sequence as well as predict the next 5 frames of images. This was compared with the actual sequence (ground truth) in order to calculate the percentage test accuracy. The last row shows the effect of randomizing the learned weights. }
    \label{simulationfigure}
\end{figure*}

\textbf{Moving MNIST digits}: This dataset consists of videos of MNIST digits. Each video was $19$ frames long and consisted  of  two digits  moving  inside  a  $64 \times 64$  patch. The digits were chosen randomly from the training set and placed initially at random locations inside the patch.  As depicted in \cite{srivastava2015unsupervised}, each digit was assigned a velocity whose direction was chosen uniformly randomly on a unit circle and whose magnitude was also chosen uniformly at random over a fixed range. The digits bounced-off the edges of the $64 \times 64$ patch and overlapped if they were at the same location. In order to reduce the learning time complexity of the task but preserve spatial complexity, we reduced the resolution of the original image patches to $16 \times 16$ bits (see top panels of Fig. 5). This makes it considerably difficult to recognize the original digits, but the patterns move in between frames in the same temporal order. We binarized the image patches using a RGB threshold value of $127$. DyBM was trained on $100$ sample videos selected randomly from the original dataset\footnote{The Moving MNIST dataset is available from \url{http://www.cs.utoronto.ca/~nitish/unsupervised_video/}.}, and then tested using another randomly selected $50$ samples. Each training video sample was reshaped into a $256 \times 15$ matrix, consisting of the first $15$ frames as input sequence to DyBM. Each column of the matrix represents the $16 \times 16$ image patch. As such the number of neurons in DyBM was set as $N = 256$, in order to encode each bit of the image. All parameters were initialized identical to the previous experiment. 

\begin{figure} 
 \centering
 \includegraphics[width =\linewidth]{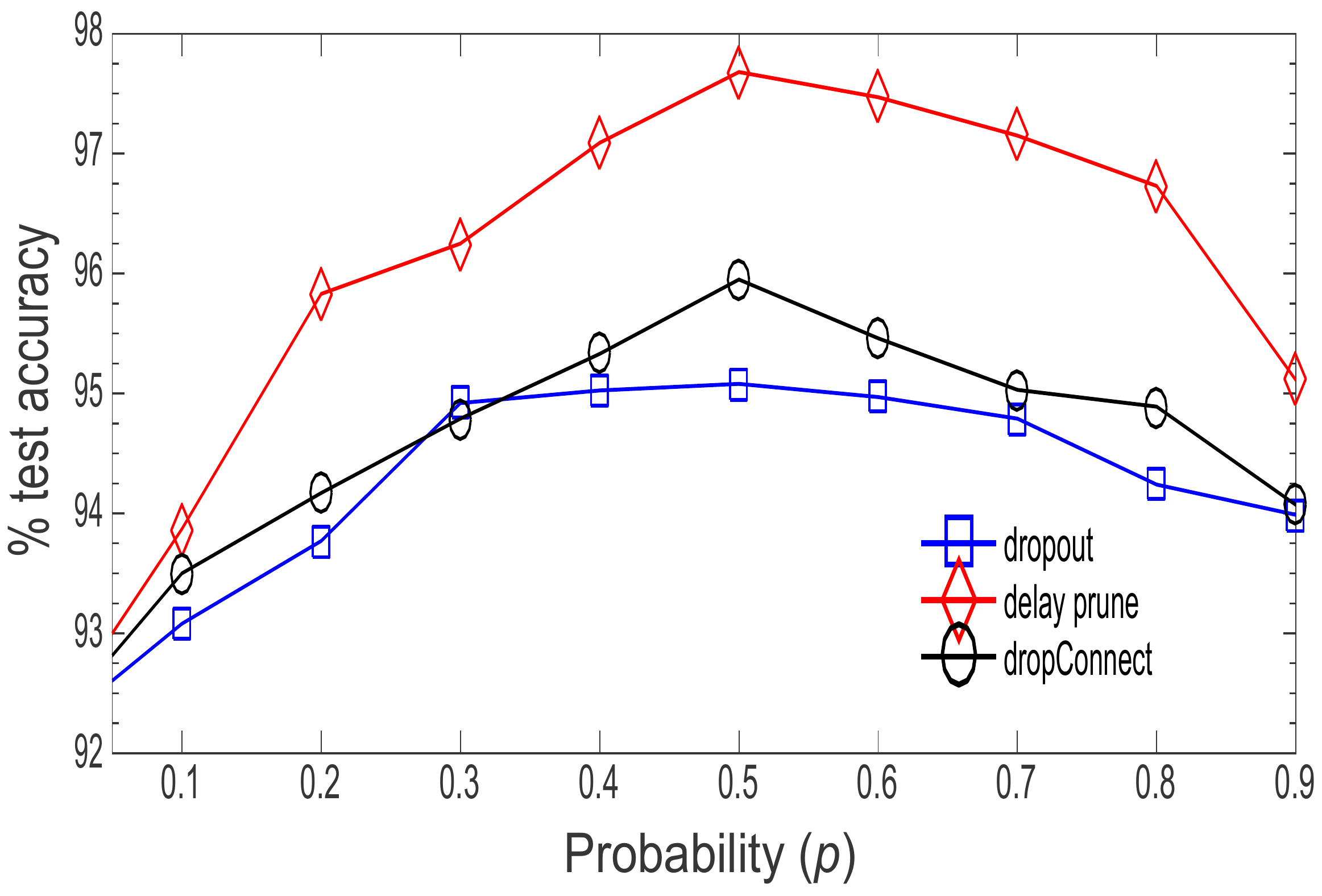}
    \caption{Performance comparison of DyBM regularized with delay pruning against regularization with Dropout and DropConnect, respectively, for the moving MNIST image sequence prediction task.}
    \label{simulationfigure}
\end{figure}

Unlike in \cite{srivastava2015unsupervised}, we used a single, considerably smaller DyBM network to learn to reconstruct the original input sequences as well as to predict into the future. Here reconstruction was tested by letting the DyBM trained on the input sequences to run forward in time up to first $15$ frames. As observed from  Fig. 5, DyBM with Delay Pruning did a significantly good job in not only reconstructing the original  $15$ input sequences but was also able to predict the next $5$ frames. Despite the relatively small training dataset, the best case test prediction accuracy as compared with the ground truth was significantly high at $97.47 \%$, for the DyBM with Delay Pruning (with a probability of $0.5$). Baseline performance of the standard DyBM (without delay pruning) was at $92.35 \%$.  As observed from the the bottom panels of Fig. 5, randomizing the learned weights completely destroyed the ability of the network to either reconstruct or predict the sequences. Prediction beyond 5 frames into the future got considerably worse, with error starting to accumulate after the predicted third frame. 

In order to compare the performance of Delay Pruning against other state of the art regularization techniques, we trained DyBM with Dropout and Dropconnect on the same task. It should be noted that, due to the peculiarity of the structure of DyBM (absence of hidden units), straightforward application of Dropout and DropConnect is difficult. In this case, we apply these techniques by considering the time unfolded DyBM-$T$, with regularisation being applied for units or connections in all layers except the units in the $0$-th layer. This layer acts analogous to the visible layer in standard RBMs. From Fig. 6 we see that the probability of deletion or pruning ($p$) effects the test prediction accuracy in all cases. However, DyBM with Delay Pruning significantly outperformed both DropConnect and Dropout regularization techniques. We thus confirmed that Delay Pruning allows robust unsupervised modelling of the video frame sequences. 

\section{Conclusion}
We have demonstrated a novel regularization technique called Delay Pruning for the Dynamic Boltzmann Machine, specially suitable for learning a generative model of multi-dimensional temporal pattern sequences. Even in the presence of a relatively small training and test dataset, Delay Pruning prevents over-fitting to give good generalized performance. Due to the uniqueness of the structure of DyBM, Delay Pruning, in the form of randomly truncating the length of  FIFO queues, leads to change in the spiking dynamics of the network by shortening the memory of spikes from a pre-synaptic to post-synaptic neuron. Experimental results show that Delay Pruning significantly outperforms other state of the art methods, enabling a 256 unit DyBM network to give a prediction accuracy of $97.47\%$ on the reduced moving MNIST dataset.\\

\textbf{Acknowledgement:} This work was supported by CREST, JST.



%

\bibliographystyle{IEEEtran}
\bibliography{AINSRef}

\end{document}